\documentclass[11pt,a4paper]{article}
\usepackage[hyperref]{acl2020}
\usepackage{times}
\usepackage{latexsym}

\usepackage{microtype}

\usepackage{graphicx} 
\usepackage{float}
\usepackage{mathtools}
\usepackage{amsmath}
\usepackage{pgfplots}
\pgfplotsset{compat=newest}
\usepgfplotslibrary{groupplots}
\usepgfplotslibrary{dateplot}

\usepackage{pdflscape}
\usepackage{makecell}
\usepackage{tabularx}

\usepackage{xcolor}

\makeatletter
\newcommand*{\centerfloat}{%
  \parindent \z@
  \leftskip \z@ \@plus 1fil \@minus \marginparwidth
  \rightskip \leftskip
  \parfillskip \z@skip}
\makeatother

\aclfinalcopy 

\title{Directed Beam Search: Plug-and-Play Lexically Constrained Language Generation}

\author{Dami\'an Pascual, B\'eni Egressy, Florian Bolli and Roger Wattenhofer \\
Department of Electrical Engineering and Information Technology \\
ETH Zurich, Switzerland \\
\{dpascual,begressy,bollif,wattenhofer\}@ethz.ch
}

\date{}

\begin{document}
\maketitle
\begin{abstract}
Large pre-trained language models are capable of generating realistic text. However, controlling these models so that the generated text satisfies lexical constraints, i.e., contains specific words, is a challenging problem. Given that state-of-the-art language models are too large to be trained from scratch in a manageable time, it is desirable to control these models without re-training them. Methods capable of doing this are called plug-and-play. Recent plug-and-play methods have been successful in constraining small bidirectional language models as well as forward models in tasks with a restricted search space, e.g., machine translation. However, controlling large transformer-based models to meet lexical constraints without re-training them remains a challenge. In this work, we propose Directed Beam Search (DBS), a plug-and-play method for lexically constrained language generation. Our method can be applied to any language model, is easy to implement and can be used for general language generation. In our experiments we use DBS to control GPT-2. We demonstrate its performance on keyword-to-phrase generation and we obtain comparable results as a state-of-the-art non-plug-and-play model for lexically constrained story generation.
\end{abstract}

\section{Introduction\footnote{Preprint. Work in progress.}}

Language generation is the task of automatically generating human-like text. This task has seen an extraordinary surge in progress since the introduction of the Transformer architecture~\cite{vaswani2017attention}, which has become the state-of-the-art architecture for language generation. Language generation is a necessary component in many different applications, such as abstractive summarization~\cite{see2017get}, dialogue generation~\cite{serban2016building}, question answering~\cite{yin2016neural} and story-telling~\cite{fan2018hierarchical}.  
For many applications, the generated text has to satisfy certain constraints. These constraints can be hard constraints, e.g., certain words must appear in the text; or soft constraints, e.g., the text should convey a certain meaning or mood. Generating natural language with hard constraints is a challenging problem, since it produces a tension between the constraint, e.g., the enforced appearance of a word in the text, and the fluency of the text. Two families of methods exist that deal with this challenge: stochastic search and enhanced beam search. 

Recent work on stochastic search~\cite{miao2019cgmh, sha2020gradient} has shown promising results on keyword to phrase generation. However, so far these methods have been applied to bidirectional Recurrent Neural Network (RNN) models and it is not clear how to adapt them to forward generation models. This is an important shortcoming since state-of-the-art language generation is dominated by 
models that 
generate language only in the forward direction~\cite{radford2019language,shoeybi2019megatron,gpt3}.
Existing enhanced beam search methods~\cite{hokamp2017lexically, post2018fast} can be directly used in forward generation models. These methods have been successful in applications where the number of candidate sequences is small, specifically in machine translation. However, current beam search methods do not have any mechanism to restrict the search space and thus, they are too expensive for general language generation. 

In this work, we present a beam search method that directs language generation towards the lexical constraints, effectively reducing the search space. Our method, which we call Directed Beam Search (DBS), encourages the language model to generate words that are similar to a given target word. Then, beam search is used to generate candidate sequences that are ranked based on the occurrence of the target word and the sequence's fluency. Our method can be combined with any language model, does not require any training or fine-tuning and is fast and easy to implement.\footnote{Code available at: \url{https://github.com/dapascual/DirectedBeamSearch}}

\section{Related Work}


Recent work has approached the problem of controllable language generation by re-training or fine-tuning a given language model. \cite{ziegler2019fine} fine-tune existing models with Reinforcement Learning, \cite{yu2017seqgan} train a Generative Adversarial Network and~\cite{keskar2019ctrl} train a large language model with 50 control codes. Different from these works, \cite{dathathri2019plug} and ~\cite{krause2020gedi} propose plug-and-play methods that can be used on state-of-the-art models without the need for additional training. The plug-and-play capability is of great importance given that current state-of-the-art language models are so large that it is unfeasible to train them from scratch each time~\cite{gpt3,raffel2019exploring,shoeybi2019megatron}. All of these methods address the problem of controlling language generation with soft constraints. 

For the case of hard constraints, \cite{xu2020megatron} recently proposed a framework for controllable language generation with hard constraints. However, this framework is not plug-and-play; it requires training three large transformer models, which is very resource and time intensive.
Two families of plug-and-play methods exist for controllable text generation with hard constrains: stochastic search and enhanced beam search. Recent stochastic search methods~\cite{miao2019cgmh, sha2020gradient} have shown their ability to produce fluent text that meets the constraints. They perform bidirectional search on the text as it is being generated and edit it until the constraints are fulfilled. These methods are suitable for bidirectional RNN models, but it is not yet clear if they can be successfully applied to forward generation models, in particular to transformer-based models. On the other hand, enhanced beam search~\cite{hokamp2017lexically} methods can be used with forward generation models. They generate a large number of candidate sequences and select the ones that meet the lexical constraints. Although improvements in computing efficiency have been proposed,~\cite{hu2019improved, post2018fast} these methods are only effective when the number of candidate sequences is small (e.g., machine translation or paraphrasing). Therefore, current enhanced beam search methods are not practical for general language generation.

Our proposed method, Directed Beam Search, permits controllable language generation with hard constraints, is plug-and-play, suitable for transformer models and can be used for general language generation. 

\section{Background}

\paragraph{Language generation} Given a sequence of input tokens $S=\{s_1, s_2, \ldots, s_m\}$, the task of a language model is to predict the next token in the sequence $s_{m+1}$. To this end, the language model produces a probability distribution over all the tokens in its vocabulary $V$ of size $|V|$. This distribution can be expressed as a vector of probabilities $\Vec{p} = \{p_1, p_2, ..., p_{|V|} \}$, where $p_i$ corresponds to the conditional probability of token $t_i$ coming next in the sequence, i.e. $p_i = p(s_{m+1} = t_i \mid s_1, s_2, \ldots, s_m)$. This vector of probabilities is obtained after applying a softmax operator oon the vector of logits $\Vec{l}$ computed by the model. State-of-the-art transformer models~\cite{radford2019language, gpt3} generate language autoregressively, i.e., at each iteration, the input of the model is the output from the previous iteration. 

\paragraph{Word embeddings} Certain algorithms for embedding words into vector representations are capable of preserving the semantic relationships between words. This means that vectors corresponding to words that have a similar meaning, are close to each other in the embedding space in terms of cosine similarity~\cite{pennington2014glove}. Several such embedding algorithms exist, such as word2vec~\cite{word2vec} or GloVE~\cite{pennington2014glove}.

\section{Directed Beam Search}

In this Section we present in detail DBS, our method for lexically constrained language generation. Given an ordered list of $n$ guide words, a language model is directed towards generating text that contains those words; the guidance is done for each of the $n$ words, in order and one at a time. Our method is agnostic of the language model used to generate the text and does not require either training or fine-tuning of the language generator or of an external discriminator. Generally speaking, DBS first increases the values of the logits of tokens that are similar to the target word. Using these modified logits, the next predicted token is sampled and the resulting sequence is auto-regressively fed back into the model. Exploiting the fact that the process of sampling the next token is stochastic, our method uses beam search to generate different candidate sequences at each iteration, which are then ranked according to a quality score. This way, our method can be divided in three stages: logit modification, beam generation and quality-based selection.

\subsection{Logit modification}

To guide language generation, DBS modifies the vector of logits $\Vec{l}$ produced by the language model (LM) right before this vector is passed through the softmax function that gives the vector of probabilities $\Vec{p}$. This modification is proportional to the cosine similarity between the target word and each word in the LM vocabulary $V$. 
For the cosine similarity between words to be meaningful, we need to operate in a space where semantic relationships exist and are measurable, such as word2vec, GloVe, etc. Without loss of generality here we use 300-dimensional GloVE embeddings~\cite{pennington2014glove}. Note that transformer-based models generate contextual representations from input tokens. Neither the space of input tokens nor the space of contextual representations describe semantic relationships and thus, they are not adequate for estimating proximity between words.

\begin{figure}[t]
    \centering
    \includegraphics[width=\linewidth]{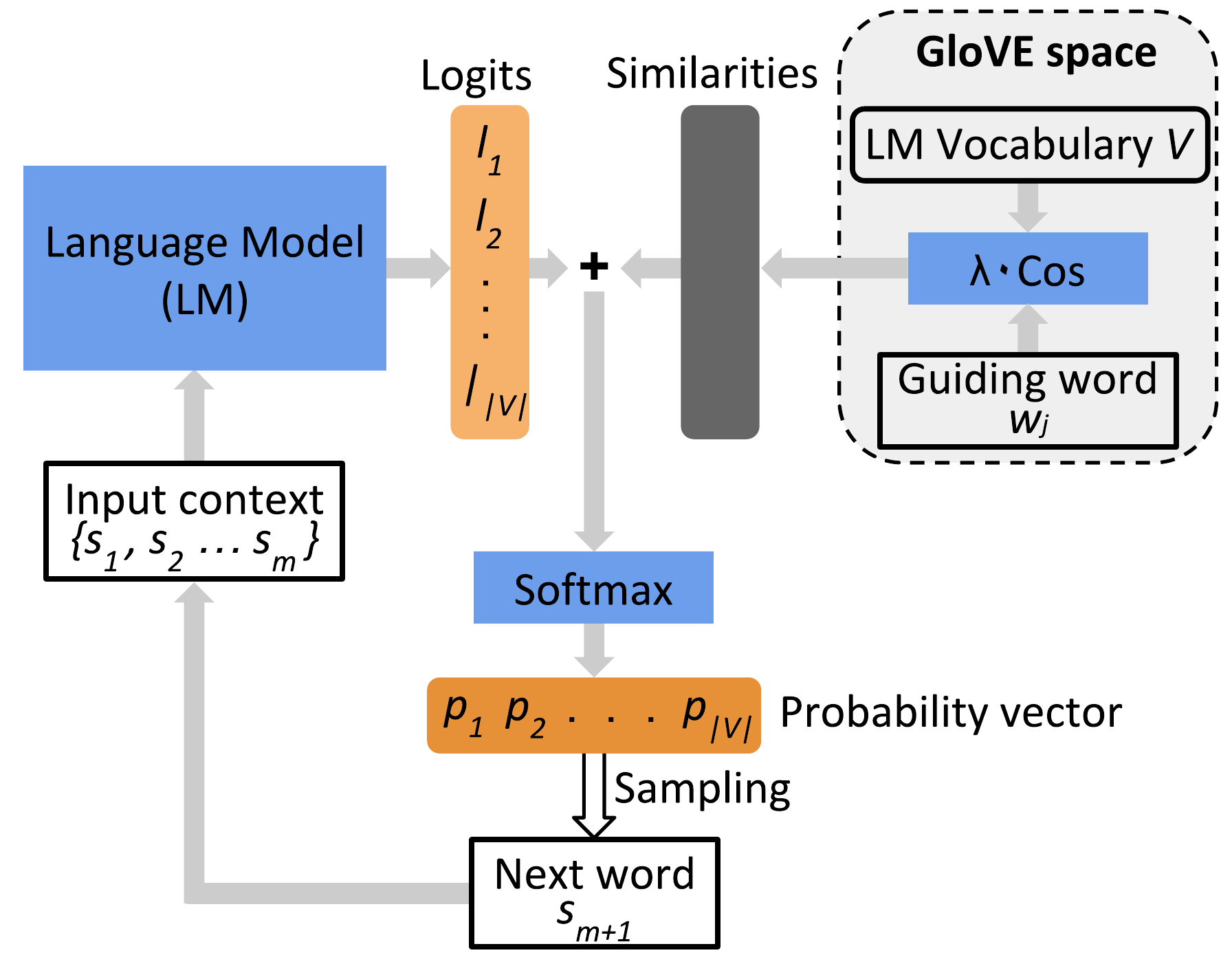}
    \caption{Logit modification mechanism.}
    \label{fig:logitMod}
\end{figure}

To adjust the logits of each word in the LM vocabulary $V$, we calculate the cosine similarity in the \emph{GloVE space} between the guide word and all the words in $V$ (not in the GloVE vocabulary, which tends to be much larger). It may occur that some tokens in the LM vocabulary do not exist in GloVE (this typically happens for tokens that do not represent a complete word), to handle this we set the GloVE emebeddings for these tokens to $\Vec{0}$ by default.
The cosine similarity can be any value between $-1$ and $1$, which we clip to be between $0$ and $1$ to avoid discouraging words that are not related to the guide word.
Then, we square the cosine similarity to sharpen the difference between words that are very similar to the guide word (similarity close to $1$) and the rest.
All in all, the adjusted logits $\Vec{l'}$ are calculated as:
$$l'_i = l_i + \lambda \cdot \min \left( 0, \cos \left(\gamma(t_i), \gamma(w_j) \right) \right)^2,$$
where $t_i$ is the i\textsuperscript{th} token of the LM vocabulary $V$, $\gamma$ denotes the GloVE embedding and $\lambda$ is a hyperparameter controlling how heavily the logits are influenced by the guide word $w_j$. An illustration of the logit modification mechanism is shown in Figure~\ref{fig:logitMod}.

\subsection{Beam Search}

As said above, $\Vec{l'}$ is the input of a softmax layer that outputs the vector of probabilities $\Vec{p'}$, from which the next word is sampled. In order to improve the quality of the text, our method uses a variant of beam search to generate candidate sequences. In the first step $b$ sequences (beams) of $k$ tokens are generated. In the following steps, for each of the $b$ beams, $s$ candidate follow-up sequences are generated, each of them consisting of $k$ tokens. A quality score (presented in the next section) is calculated for all the $b \times s$ sequences and the best $b$ sequences (beams) are kept. The process is repeated until a termination condition is reached, e.g., maximum number of tokens to generate. When the j\textsuperscript{th} guide word $w_j$ appears, guidance, i.e., logits modification, is immediately disabled and the rest of the $k$ tokens in the current sequence are generated without guidance. This way, the sequence length $k$ can also be used to softly control the distance between guide words in the generated text.  

After a guide word occurs in one beam, that beam switches guidance to the next word $w_{j+1}$. We use word stemming to check for the occurrence of the guide words, i.e., if a generated word has the same stem as the guide word, it is counted as an occurrence; this way special cases such as plurals can be automatically handled, whilst keeping the text coherent by avoiding semantic redundancies, e.g., ``protective protection''. 

All in all, our variant of beam search has three hyperparameters, $b$, $s$ and $k$. The first two, $b$ and $s$, determine the number of beams and candidate sequences, which in turn regulates the trade-off between text quality and time required to generate the final sequence. As mentioned, $k$ is related to the density of the guide words in the generated text.



\subsection{Quality Score}

Our method is based on a quality score that rewards the appearance of the guide word as well as sequences with low perplexity scores~\cite{chen1998evaluation} (perplexity is a measure of fluency, lower values represent better fluency). In order to prevent repetitions, the score penalizes sequences with multiple occurrences of the guide word. Formally, the quality score $Q$ of a sequence $S$ is given by:

\begin{equation*}
    Q(S) = 
    \begin{cases}
    \exp \left( -(c + \alpha PP(S)) \right) &\text{if $c > 0$}\\
    \exp \left( -(c^* + \alpha PP(S)) \right) &\text{else}
    \end{cases}
\end{equation*}

\begin{figure}
    \centering
    \includegraphics[width=\linewidth]{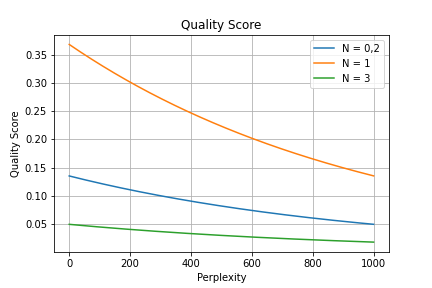}
    \caption{Quality score for different numbers of word occurrences and perplexities with hyperparameters $c^* = 2$ and $\alpha = 0.001$.}
    \label{fig:quality_score}
\end{figure}

Here, $c$ is a counter of the number of occurrences of the guide word in the predicted sequence, $PP(S)$ is the perplexity of sequence $S$, $\alpha$ is a hyperparameter controlling the relative importance of the word count and the perplexity score, and $c^*$ is a hyperparameter penalizing sequences with no appearance of the guide word. We use $c^* = 2$, which implies we favor $c = 1$ over $c = 0$. And we use $\alpha=0.001$, which implies we prioritise perplexity over number of occurrences only if the perplexity is very high ($>1000$). See Figure \ref{fig:quality_score} 
for a comparison of the quality score for different values of $c$ and $PP$. As before, stemming is used to detect word occurrences. The candidate beams generated in the beam search step are sorted by their cumulative quality score, i.e., the sum of the quality scores of all the (sub)sequences that constitute the beam. 

This way, DBS directs beam search towards lexically constrained text in two ways: 1) by increasing the probability of sampling the words semantically similar to the guide word; and 2) by selecting candidate sequences that contain the guide word, while maintaining good fluency, i.e., low perplexity. We emphasize that this is a general approach that can be used out-of-the-box for directing any forward language generation model.

\section{Experimental Setup}

To evaluate our method, we run two sets of experiments. First, we evaluate with objective metrics the quality of the text generated by DBS in the task of keyword-to-phrase generation. We use these results to study the effect of the different hyperparameters. Then, we use our method in keyword-guided story generation and compare some examples of the lexically constrained text generated by DBS with another state-of-the-art methods. In all of our experiments we use GPT-2 large ($774$ Million parameters)~\cite{radford2019language} as our language model with top-p sampling ($p=0.9$)~\cite{holtzman2019curious} and temperature equals $1$. 

\subsection{Keyword to Phrase Generation}


This task consists of generating a piece of text that contains certain keywords. To select the keywords we use a list of $1000$ common English words\footnote{\url{https://github.com/powerlanguage/word-lists/}} and discard the first $500$ words (these words are so common that it would be hard to say if the word appeared thanks to DBS or not). From the remaining $500$, we discard stop words and then we sample with replacement $50$ different sets of $5$ words each. For each of these sets, we calculate \emph{success rate}, \emph{perplexity} and \emph{success length}:

\begin{itemize}
    \item \textbf{Success rate:} Number of keywords that appear in the text divided by the total number of keywords, in this case $5$.
    \item \textbf{Perplexity:} Measure of the fluency of the text, lower is better. Calculating perplexity with the same model that generates the text is problematic~\cite{liu2016not}, therefore, we use ``Distil-GPT-2''~\cite{wolf2019huggingface} to calculate the perplexity of the generated text.
    \item \textbf{Success length:} Number of tokens generated until all the keywords occur. If all the keywords do not appear before termination, the success length is set to the total number of tokens generated.
\end{itemize}

In this experiment we use as initial context for GPT-2 the phrase ``It is''. Starting with this phrase the model generates a piece of text directed by DBS towards the keywords. We set termination to occur after $90$ tokens have been generated. We use this experiment to study the behavior of the method under different combinations of hyperparameters.

The most important hyperparameter we need to determine is the guidance strength $\lambda$. Additionally, we need to set the number of beams $b$ and the number of concurrent sequences $s$. The number of words per sequence $k$ helps us regulate the density of keywords, i.e., the success length. 

\subsection{Story Generation}

Next, we informally compare our method against the \textsc{Megatron-CTRL} model~\cite{shoeybi2019megatron} in story generation on the ROC dataset~\cite{mostafazadeh2016corpus}. The \textsc{Megatron-CTRL} model is not plug-and-play and requires an expensive training phase. This model uses a knowledge base to keep the coherence of the generated text and a GPT-2 model to generate text. We compare DBS on the 774 million parameter GPT-2 against the \textsc{Megatron-CTRL-8B}, which uses a GPT-2 model with 8 billion parameters to generate text. 

In this experiment we generate controlled text for the same examples and keywords as presented in~\cite{shoeybi2019megatron}. We defer a more formal comparison to a later version of this work.

\section{Results}

\subsection{Keyword to Phrase Generation}

\begin{figure}[t]
\centerfloat
\begin{tikzpicture}

\definecolor{color0}{rgb}{0.12156862745098,0.466666666666667,0.705882352941177}
\definecolor{color1}{rgb}{1,0.549019607843137,0}

\begin{axis}[
tick align=outside,
tick pos=left,
x grid style={white!69.0196078431373!black},
xmin=-0.585, xmax=4.585,
xtick style={color=black},
xtick={0,1,2,3,4},
xticklabels={5,10,15,20,25},
y grid style={white!69.0196078431373!black},
ylabel={Success rate},
ymin=0, ymax=1.05,
ytick style={color=black},
ytick={0,0.2,0.4,0.6,0.8,1,1.2},
yticklabels={0.0,0.2,0.4,0.6,0.8,1.0,1.2},
width = 0.98\linewidth,
tick label style={font=\small},
label style={font=\small}
]
\draw[draw=none,fill=color0] (axis cs:-0.35,0) rectangle (axis cs:0,0.32);
\draw[draw=none,fill=color0] (axis cs:0.65,0) rectangle (axis cs:1,0.74);
\draw[draw=none,fill=color0] (axis cs:1.65,0) rectangle (axis cs:2,0.84);
\draw[draw=none,fill=color0] (axis cs:2.65,0) rectangle (axis cs:3,0.86);
\draw[draw=none,fill=color0] (axis cs:3.65,0) rectangle (axis cs:4,0.84);
\draw (axis cs:-0.245,0.32) ++(0pt,1pt) node[
  scale=0.8,
  anchor=south,
  text=black,
  rotate=0.0
]{0.32};
\draw (axis cs:0.825,0.74) ++(0pt,1pt) node[
  scale=0.8,
  anchor=south,
  text=black,
  rotate=0.0
]{0.74};
\draw (axis cs:1.825,0.84) ++(0pt,1pt) node[
  scale=0.8,
  anchor=south,
  text=black,
  rotate=0.0
]{0.84};
\draw (axis cs:2.825,0.86) ++(0pt,1pt) node[
  scale=0.8,
  anchor=south,
  text=black,
  rotate=0.0
]{0.86};
\draw (axis cs:3.825,0.84) ++(0pt,1pt) node[
  scale=0.8,
  anchor=south,
  text=black,
  rotate=0.0
]{0.84};
\end{axis}

\begin{axis}[
axis y line=right,
legend cell align={left},
legend style={fill opacity=0.8, draw opacity=1, text opacity=1, at={(0.01,0.99)}, nodes={scale=0.8}, anchor=north west, draw=white!80!black},
tick align=outside,
x grid style={white!69.0196078431373!black},
xmin=-0.585, xmax=4.585,
xtick pos=left,
xtick style={color=black},
xticklabels={},
y grid style={white!69.0196078431373!black},
ylabel={Perplexity},
ymin=0, ymax=26,
ytick pos=right,
ytick style={color=black},
width = 0.98\linewidth,
tick label style={font=\small},
label style={font=\small}
]

\addlegendimage{ybar,ybar legend,draw=none,fill=color0};
\addlegendentry{Success Rate}
\addlegendimage{ybar,ybar legend,draw=none,fill=color1};
\addlegendentry{Perplexity}

\draw[draw=none,fill=color1] (axis cs:2.77555756156289e-17,0) rectangle (axis cs:0.35,12.4);
\draw[draw=none,fill=color1] (axis cs:1,0) rectangle (axis cs:1.35,15.39);
\draw[draw=none,fill=color1] (axis cs:2,0) rectangle (axis cs:2.35,16.6);
\draw[draw=none,fill=color1] (axis cs:3,0) rectangle (axis cs:3.35,16.03);
\draw[draw=none,fill=color1] (axis cs:4,0) rectangle (axis cs:4.35,14.2);
\end{axis}

\end{tikzpicture}
\caption{Comparison of success rate and perplexity for different values of $\lambda$ with $b=s=c=5$.}
\label{lambda}
\end{figure}
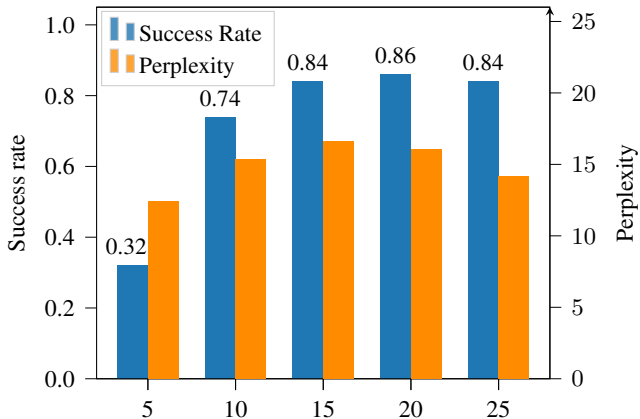

\begin{figure*}[t]
    \centering
    \includegraphics[trim={0 1cm 0 0},clip,width=\linewidth]{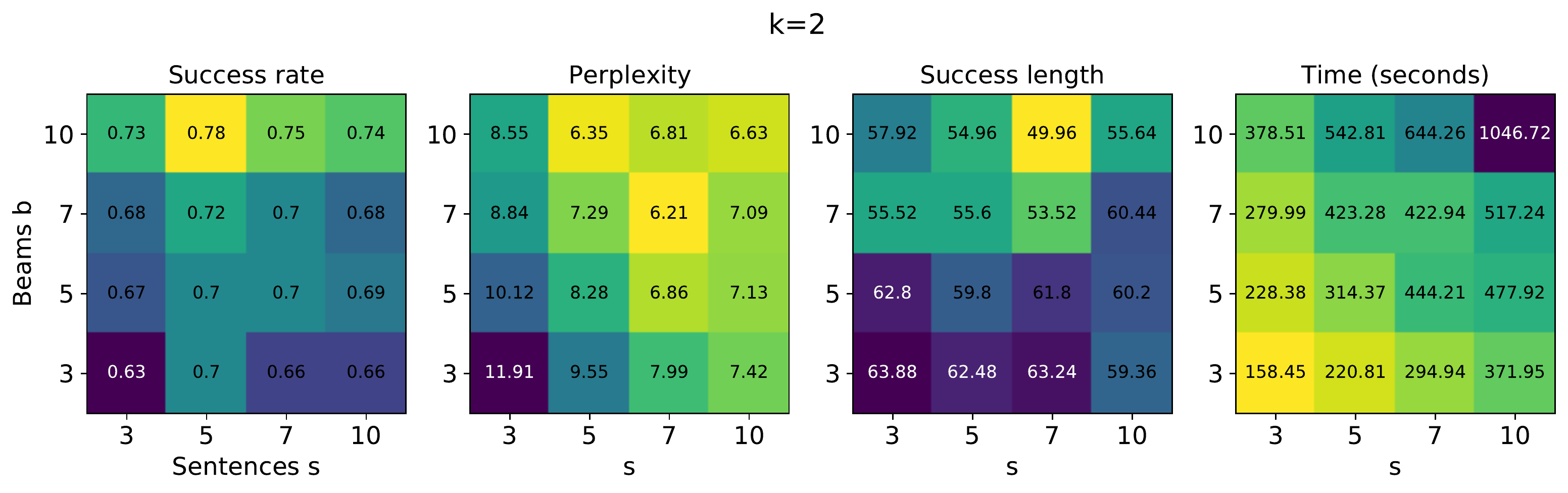}
    \caption{Keyword-to-phrase results for $k=2$.}
    \label{fig:k2}
\end{figure*}
\begin{figure*}[t]
    \centering
    \includegraphics[trim={0 1cm 0 0},clip,width=\linewidth]{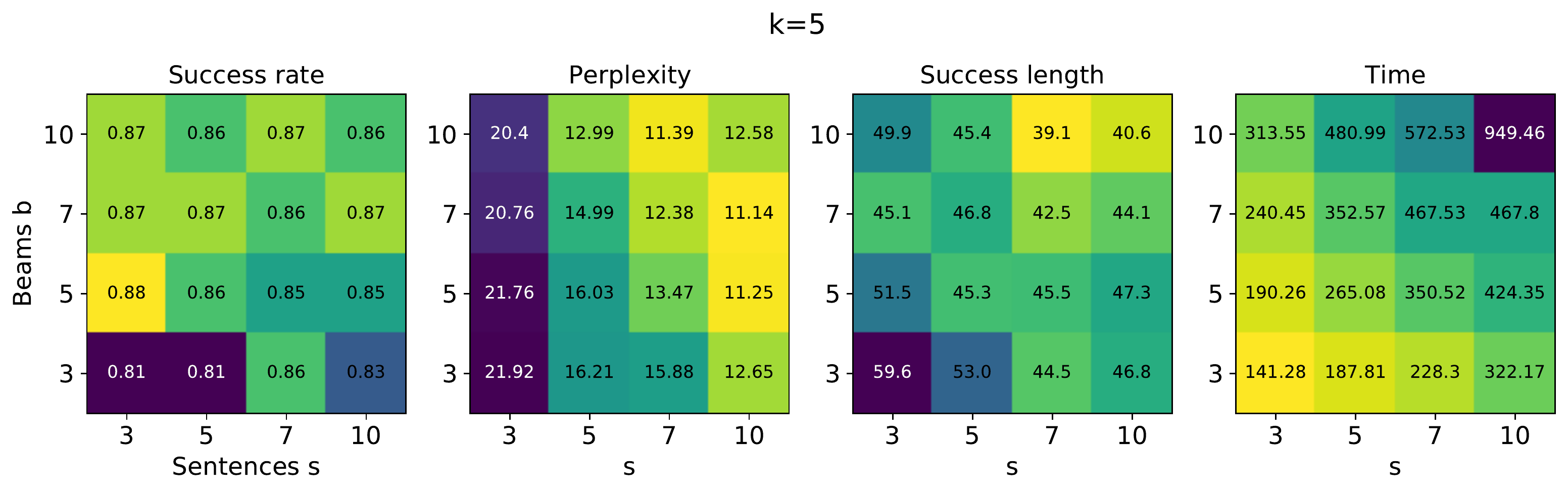}
    \caption{Keyword-to-phrase results for $k=5$.}
    \label{fig:k5}
\end{figure*}

In this experiment, we first study the impact of the guidance strength $\lambda$. To this end, we fix the number of beams $b$, the number of concurrent sequences $s$ and the tokens per sequence to $5$. For $\lambda$ we consider the values $5$, $10$, $15$, $20$ and $25$. In Figure~\ref{lambda} we report the mean success rate and perplexity across the $50$ keyword sets for each value of $\lambda$.

We see that for all values of lambda perplexity is in the same range of values: between $12$ and $17$. Lower values of $\lambda$ present lower perplexities, which is expected due to the trade-off between fulfilling the constraints and generating fluent text. The success rate is the highest for $\lambda=20$, with a mean value of $0.86$ across the $50$ sets of keywords. The success rate for $\lambda=15$ and $\lambda=25$ is $0.84$ while, as expected, the success rate quickly decreases with lower values of $\lambda$. For the rest of the experiments we pick $\lambda=20$. Since small changes in $\lambda$ do not have a strong effect on the results, we do not tune its value any further. 

Next, we study the performance of DBS when varying the number of beams $b$, the number of concurrent sequences $s$ and the sequence length $k$. For $b$ and $s$ we consider the values $3$, $5$, $7$ and $10$; for $k$ we consider $2$, $5$ and $10$. We report success rate, perplexity, success length and computing time for each combination. For reference, we run language generation without DBS and with a single beam for the $50$ keyword sets and obtain a mean perplexity of $33.3$ and a mean succes rate of $0.01$, which shows that the random baseline for the success rate is almost $0$.

In Figure~\ref{fig:k2}, we see that for $k=2$ the perplexity is as low as $6.21$. However the maximum success rate is only $0.78$ while the success length revolves around $50$ words for the best performing combinations. On the other hand, for $k=5$ (Figure~\ref{fig:k5}), while perplexity is a bit higher, between $11$ and $12$ for the best combinations, success rate stays consistently above $0.85$ and the best success length is as low as $39.1$. When $k=10$ (detailed results are shown in Appendix~\ref{sec:appendix}), the maximum success rate is $0.93$. However, both perplexity and success length are significantly worse than for $k=2$ and $k=5$. Finally, it is worth noting that as expected, a larger amount of beams and concurrent sentences results in more computing time. In particular, choosing $b$ and $s$ larger than $7$ slows down computation significantly without bringing much improvement in performance; for instance, for $k=5$, when $b=10$ and $s=10$ the required time is $3.5$ times more than when $b=5$ and $s=5$, whilst the success rate stays the same (perplexity and success length improve slightly).


\renewcommand{\arraystretch}{1.3}
\begin{table}[h]
\centering
\begin{tabularx}{\columnwidth}{X}
\hline
Keywords: \textit{Enemy, Speed, Meet, Colony, Mouth} \\
\hline
Text: \textit{It is the \textbf{enemy} that is the real problem. The \textbf{speed} with which we meet it in the \textbf{colonies} will determine the fate of the planet." A \textbf{mouthful} of spit echoed around the room.}
\\
\hline
\hline
Keywords: \textit{Cost, Thank, Summer, Control, Nose} \\
\hline
Text: \textit{It is more \textbf{cost} effective in the long run. \textbf{Thank} you again for making this year the best \textbf{summer} I have ever had at the \textbf{controls} of the \textbf{nose} cone!}\\
\hline
\hline
Keywords: \textit{Desert, Wild, Crop, Bear, Enemy } \\
\hline
Text: \textit{It is a \textbf{desert} country, and \textbf{wild} animals are plentiful. Here and there, \textbf{crops} \textbf{bear} fruit. There are \textbf{enemies} as well as friends.}\\
\hline
\hline

\end{tabularx}
\caption{Examples of text generated by DBS from keywords, the input text is ``It is''.}
\label{tab:keyword2phrase}
\end{table}
\renewcommand{\arraystretch}{1}

These results show that GPT-2 guided by DBS produces text with lower perplexity than GPT-2 without beam search, while meeting up to 93\% of the constraints on average for the most time-consuming combinations. From these results, we select $k=5$, $b=7$ and $s=10$, since this combination offers a good trade-off between performance and computing time. Table~\ref{tab:keyword2phrase} provides examples of the text generated with this combination of hyperparameters; these examples reveal the difficulty of the task due to the random selection of keywords.

\subsection{Story Generation}

Finally, we show in Table~\ref{tab:DBSvsMC} examples of text generated by \textsc{Megatron-cntrl-8B}~\cite{shoeybi2019megatron} and by DBS on \textsc{Gpt-2-774M}. These examples show that in both cases the lexical constraints are met and that the quality of the text is comparable despite the large difference in the number of parameters and in the complexity of the control methods. 

\renewcommand{\arraystretch}{1.3}
\begin{table*}[t]
\centering
\begin{tabularx}{\linewidth}{X}
\hline
Context: \textcolor{blue}{He and his friends are bored on a weekend day.} Keywords: go, decide \\
\hline
\textsc{Megatron-cntrl-8B}~\cite{shoeybi2019megatron}: \textit{They decided to go to the video game store. He picks out a new video game. He and his friends have fun playing the new video game. He is no longer bored.}\\
\hline
\textsc{DBS + Gpt2-774M}: \textit{They go for a walk and decide to go for a bike ride. As they are riding in the bike lane, a car drives up in front of them. The car stops in front of them and two men get out.} 
\\
\hline
\hline
Context: \textcolor{blue}{He had a heavy duty backpack.} Keywords: took \\
\hline
\textsc{Megatron-cntrl-8B}: \textit{He took it to the army base camp. He packed it with items like food, ammo, and tools. He took it to the jungle every day. One day, he took it out and his backpack exploded.}\\
\hline
\textsc{DBS + Gpt2-774M}: \textit{He took off his shoes and socks and laid them out in front of him on the floor. Then he pulled out a blanket and draped it over his knees.} 
\\
\hline
\hline
Context: \textcolor{blue}{He hiked up the local hill.} Keyword: top\\
\hline
\textsc{Megatron-cntrl-8B}: \textit{He was at the top of the hill looking down. the weather looked beautiful. the clouds suddenly passed. the sun began to shine over the land again.}\\
\hline
\textsc{DBS + Gpt2-774M}: \textit{The top of the hill had a clear view of the town. "Did you see that?" I asked. "Yeah."} 
\\
\hline
\hline
Context: \textcolor{blue}{He liked trading commodities. one day he was trading several products. He noticed that he gained a lot of money from the new trades.} Keyword: realize\\
\hline
\textsc{Megatron-cntrl-8B}: \textit{He realized that the new trades were his best trading idea. he quit his old job and went fulltime on his new trading idea.}\\
\hline
\textsc{DBS + Gpt2-774M}: \textit{So he decided to start trading. After a few days of trading, he realized that there was a big gap between the price of the products he was selling and the price of the products he was buying.} 
\\
\hline
\hline

\end{tabularx}
\caption{Comparison of examples generated by \textsc{Megatron-CTRL} and DBS+GPT2 for the same input context and keywords.}
\label{tab:DBSvsMC}
\end{table*}


\section{Conclusion}

In this work we have presented DBS a plug-and-play method that controls language generation in order to generate text that contains specific words. We have demonstrated that DBS can guide large state-of-the-art pre-trained models, like GPT-2, to meet lexical constraints without a degradation in perplexity. Our evaluation on keyword-to-phrase generation shows the effectiveness of DBS in terms of success rate and fluency of the text. Furthermore, we provide a detailed study of the performance of the model for different hyperparameter combinations. All in all, DBS is a simple yet effective method for lexically constrained language generation that can be applied to any language model without the need for training or fine-tuning.

\bibliography{references}
\bibliographystyle{acl_natbib}

\appendix
\onecolumn
\clearpage
\section{Evaluation results for k=10}\label{sec:appendix}

\begin{figure*}[h]
    \centering
    \includegraphics[trim={0 0 0 1cm},clip,width=\linewidth]{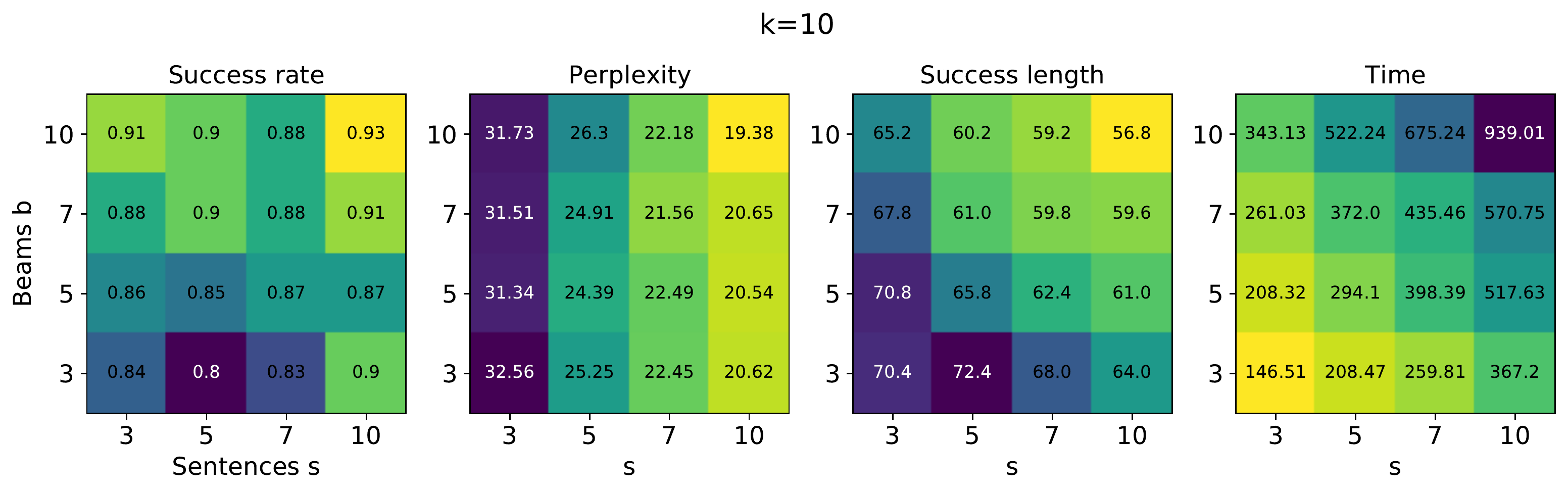}
    \caption{Keyword-to-phrase results for $k=10$.}
    \label{fig:k10}
\end{figure*}

\end{document}